\def\year{2019}\relax
\documentclass[letterpaper]{article} 
\usepackage{aaai19}  
\usepackage{times}  
\usepackage{helvet}  
\usepackage{courier}  
\usepackage{url}  
\usepackage{graphicx}  

\usepackage{amsmath}
\usepackage{algorithm, algorithmicx, algpseudocode}
\usepackage{multirow,array}
\usepackage{enumitem}

\newcommand{\citet}[1]{\citeauthor{#1}~\shortcite{#1}}

\begin{document}
%
\title{Style Transfer as Unsupervised Machine Translation}

\author{ Zhirui Zhang$^\dag$\thanks{The first two authors contributed equally to this work.}, Shuo Ren$^\ddag$, Shujie Liu$^\S$, Jianyong Wang$^\P$, Peng Chen$^\P$, \\ {\bf \Large Mu Li$^\natural$, Ming Zhou$^\S$, Enhong Chen$^\dag$} \\
  $^\dag$University of Science and Technology of China, Hefei, China\\
  $^\ddag$SKLSDE Lab, Beihang University, Beijing, China \\ 
   $^\S$Microsoft Research Asia \ $^\P$Microsoft Research and AI Group \\
    $^\dag$zrustc11@gmail.com\ \ cheneh@ustc.edu.cn\ \ $^\ddag$shuoren@buaa.edu.cn \\ 
    $^{\S\P}$\{shujliu,peche,mingzhou\}@microsoft.com\ $^{\natural}$limugx@outlook.com \\
}
\maketitle
\begin{abstract}
Language style transferring rephrases text with specific stylistic attributes while preserving the original attribute-independent content.
One main challenge in learning a style transfer system is a lack of parallel data where the source sentence is in one style and the target sentence in another style. 
With this constraint, in this paper, we adapt unsupervised machine translation methods for the task of automatic style transfer. 
We first take advantage of style-preference information and word embedding similarity to produce pseudo-parallel data with a statistical machine translation (SMT) framework. 
Then the iterative back-translation approach is employed to jointly train two neural machine translation (NMT) based transfer systems. 
To control the noise generated during joint training, a style classifier is introduced to guarantee the accuracy of style transfer and penalize bad candidates in the generated pseudo data. 
Experiments on benchmark datasets show that our proposed method outperforms previous state-of-the-art models in terms of both accuracy of style transfer and quality of input-output correspondence.

\end{abstract}

\section{Introduction}

\begin{figure}[t] 
	\begin{center}
		\includegraphics[scale=0.54]{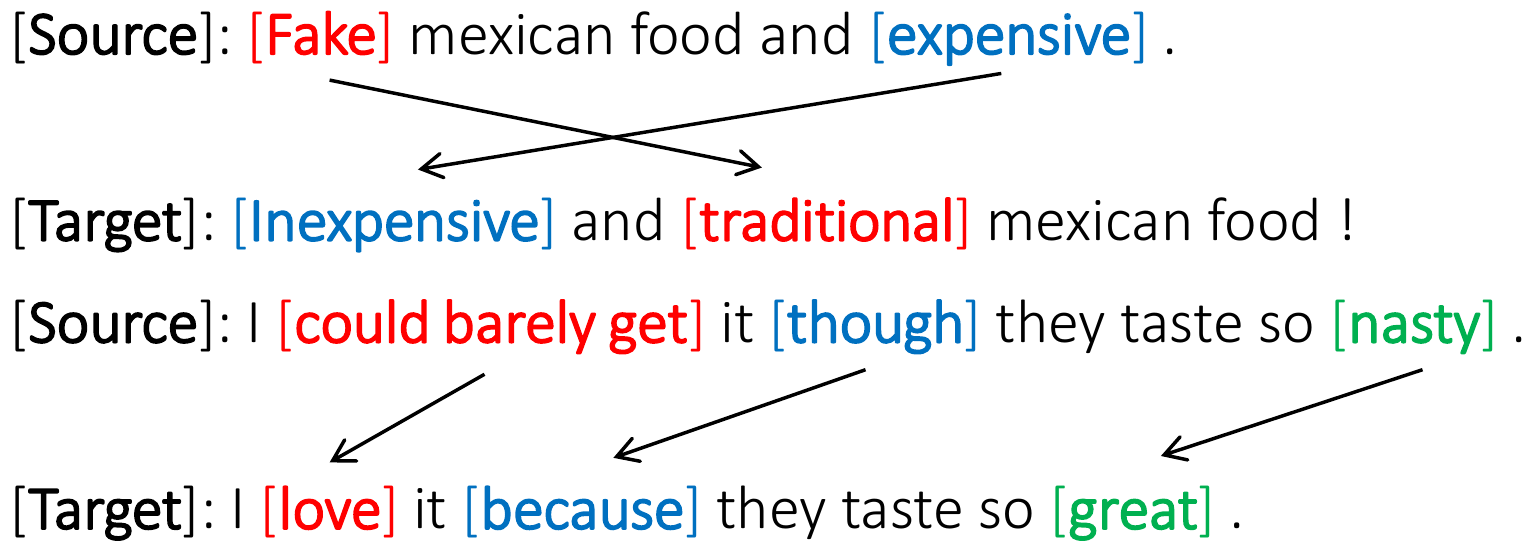}
	\end{center}
    \vspace*{-10pt}
    \caption{Some examples of language style transfer (e.g., from negative sentiment to positive sentiment). The arrow indicates the transformation of different words from the source attribute to the target attribute.}
		\label{fig:introduce-example}
\end{figure}

Language style transfer is an important component of natural language generation (NLG) \cite{Wen2015SemanticallyCL,Li2016APN,Sennrich2016ControllingPI,Wintner2017PersonalizedMT}, as it enables NLG systems to control not only the topic of produced utterance but also attributes such as sentiment and gender.
As shown in Figure \ref{fig:introduce-example}, language style transfer aims to convert a sentence with one attribute (e.g., negative sentiment) to another with a different attribute (e.g., positive sentiment), while retaining its attribute-independent content (e.g., the properties of the product being discussed).

Recently, many methods have made remarkable progress in language style transfer.
One line of research \cite{Hu2017TowardCG,Shen2017StyleTF,Fu2017StyleTI} leverages the auto-encoder framework to learn an encoder and a decoder, in which the encoder constructs a latent vector by removing the style information and extracting attribute-independent content from the input sentence, and the decoder generates the output sentence with the desired style. 
Another line involves a delete-retrieve-generate approach \cite{Li2018DeleteRG,unpaired-sentiment-translation}, in which attribute-related words are recognized and removed to generate a sentence containing only content information, which is used as a query to find a similar sentence with the target attribute from the corpus. Based on that, target attribute markers can be extracted and utilized to generate the final output sentence in a generation step.

Language style transfer can be regarded as a special machine translation (MT)  task where the source sentence is in one style and the target sentence is in another style (as shown in Figure \ref{fig:introduce-example}). 
In this paper, we leverage attention-based neural machine translation (NMT) models \cite{sutskever2014sequence,Cho2014LearningPR,Bahdanau2014NeuralMT} to change the attribute of the input sentence by translating it from one style to another.
Compared with auto-encoder methods, the attention mechanism can better make the decision on preserving the content words and transferring attribute-related words.
Compared with the delete-retrieve-generate approach, our model is an end-to-end system without error propagation, and the generation of target attribute words is generated based on the context information instead of a retrieval step.

To train NMT-based systems, a large parallel corpus is required to tune the huge parameters and learn the correspondence between input and output words.
However, for style transfer, sentence pairs with the same content but different attributes are difficult to acquire. 
Inspired by unsupervised MT approaches \cite{Artetxe2018UnsupervisedNM,Piotr2018UnsupervisedMT,lample2018phrase}, we propose a two-stage joint training method to boost a forward transfer system (source style to target style) and a backward one (target style to source style) using unpaired datasets. 
In the first stage, we build the word-to-word transfer table based on word-level style-preference information and word embedding similarity learnt from unpaired datasets. 
With the inferred transfer tables and pre-trained style specific language models, bidirectional (forward and backward) statistical machine translation (SMT) \cite{och2003minimum,chiang2007hierarchical} transfer systems are built to generate a pseudo parallel corpus.
In the second stage, we initialize bidirectional NMT-based transfer systems with the pseudo corpus from the first stage, which are then boosted with each other in an iterative back-translation framework.
During iterative training, a style classifier is introduced to guarantee the high accuracy of style transfer result and punish the bad candidates in the generated pseudo data.

We conduct experiments on three style transfer tasks: altering sentiment of Yelp reviews, altering sentiment of Amazon reviews, and altering image captions between romantic and humorous.
Both human and automatic evaluation results show that our proposed method outperforms previous state-of-the-art models in terms of both accuracy of style transfer and quality of input-output correspondence (meaning preservation and fluency).
Our contributions can be summarized as follows:
\begin{itemize}
	\item Unsupervised MT methods are adapted to the style transfer tasks to tackle the lack of parallel corpus, with a three-step pipeline containing building word transfer table, constructing SMT-based transfer systems and training NMT-based transfer systems. 
    \item Our attention-based NMT models can directly model the whole style transfer process, and the attention mechanism can better make the decision of preserving the content words and transferring the attribute-related words.
	\item A style classifier is introduced to control the noise generated during iterative back-translation training, and it is crucial to the success of our methods.
\end{itemize}

\begin{figure*}[th] 
	\begin{center}
		\includegraphics[scale=0.52]{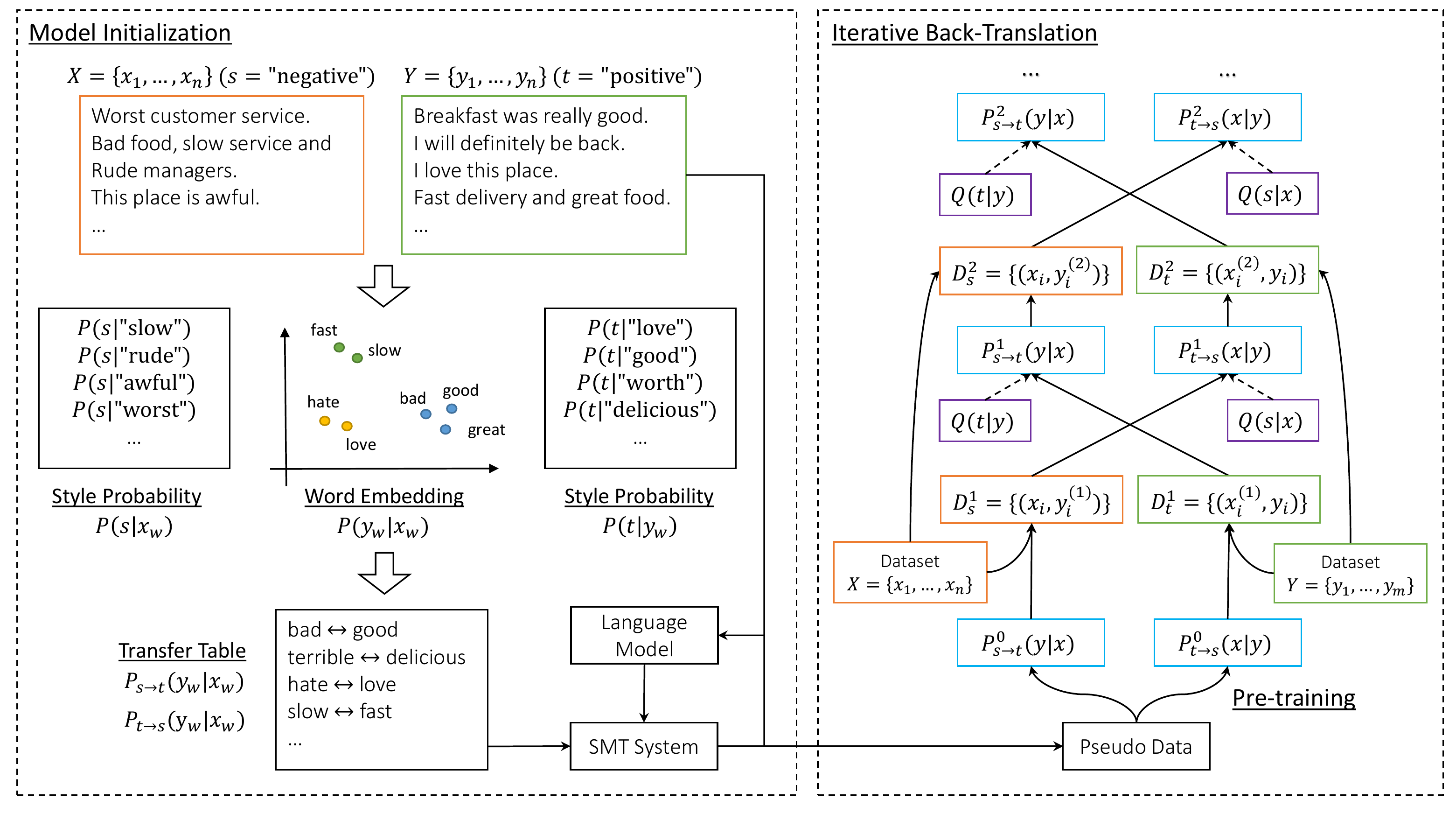}
	\end{center}
      	\vspace*{-10pt}
    	\caption{Illustration of the overall training framework of our approach. This framework consists of model initialization and iterative back-translation components, in which $P(s|x_w)$ and $P(t|y_w)$ denote style preference probabilities of words $x_w$ and $y_w$, $P(y_w|x_w)$ represents word similarity defined in the embedding space, $P_{s \rightarrow t}(y_w|x_w)$ and $P_{t \rightarrow s}(x_w|y_w)$ stand for the transfer probability of different words, $P_{s\rightarrow t}(y|x)$ and $P_{t\rightarrow s}(x|y)$ are source-to-target and target-to-source style transfer models, $Q(s|x)$ and $Q(t|y)$ denote the probabilities that a sentence belongs to different styles, and they are used to punish poor pseudo sentence pairs with wrong attributes.}
		\label{fig:model-framework}
\end{figure*}

\section{Our Approach}

Given two datasets $X=\{x_1,..., x_{n}\}$ and $Y=\{y_1,..., y_{m}\}$ representing two different styles $s$ and $t$ respectively (e.g., for the sentiment, $s=``\text{negative}"$, $t=``\text{positive}"$), style transfer can be formalized as learning the conditional distribution $P_{s\rightarrow t}(y|x)$, which takes $(x,s)$ as inputs and generates a sentence $y$ retaining the content of $x$ while expressing in the style $t$.
To model this conditional distribution, we adopt the attention-based architecture proposed by~\citet{Bahdanau2014NeuralMT}.
It is implemented as an encoder-decoder framework with recurrent neural networks (RNN), in which RNN is usually implemented as Gated Recurrent Unit (GRU)~\cite{Cho2014LearningPR} or Long Short-Term Memory (LSTM) networks~\cite{Hochreiter1997LongSM}.
In our experiment, GRU is used as our RNN unit.

To learn style transfer using non-parallel text, we design an unsupervised sequence-to-sequence training method as illustrated in Figure \ref{fig:model-framework}.
In general, our proposed approach can be divided into two stages: model initialization and iterative back-translation. 
In the first stage, given unaligned sentences $X=\{x_1,..., x_{n}\}$ and $Y=\{y_1,..., y_{m}\}$, we first build the transfer table to provide word-to-word transfer information, as well as two style specific language models. 
With the word-to-word transfer table and language models, we build two SMT-based transfer systems (source-to-target model and target-to-source model), with which we translate the unaligned sentences to construct the pseudo-parallel corpus.
In the second stage, we use the pseudo data to pre-train bidirectional NMT-based transfer systems (source-to-target model $P_{s\rightarrow t}(y|x)$ and target-to-source model $P_{t\rightarrow s}(x|y)$).
Based on the two initial systems, an iterative back-translation algorithm is employed to sufficiently exploit unaligned sentences $X$ and $Y$, with which bidirectional systems can achieve further improvements.

\subsection{Model Initialization}

Learning to style transfer  with only non-parallel data is a challenging task, since the associated style expressions cannot be learnt directly.
To reduce the complexity of this task, we first learn the transfer knowledge at the word level, with which we can upgrade to the sentence level. 
To achieve this goal, we first construct word-level transfer table $P_{s\rightarrow t}(y_w|x_w)$ in an unsupervised way. 
Many methods \cite{Conneau2017WordTW,Artetxe2017LearningBW} have been proposed to perform a similar task, but these methods rely on the homogeneity of the cross-lingual word embedding space and are only applied in the MT field.
Since the two style transfer corpora are in one language, cross-lingual word embedding cannot be used to learn word-level transfer information.
In order to gain proper word mapping between different attributes, we propose a new method which leverages the word embedding similarity and style preference of  words to construct word-level transfer table.

The transfer probability $P_{s\rightarrow t}(y_w|x_w)$ between source word $x_w$ in style $s$ and target word $y_w$ in style $t$ can be decomposed into three parts:
\begin{equation}
\begin{aligned}
 P_{s \rightarrow t}(y_w|x_w) & = P(y_w|x_w, s, t) = \frac{P(y_w,s,t|x_w)}{P(s,t)}  \\
 		 = & \frac{ P(s|x_w) P(y_w|x_w,s)P(t|x_w,y_w,s)}{P(s,t)} \\
         \propto & \ P(s|x_w)P(y_w|x_w)P(t|y_w)
\end{aligned}
\label{equ:trans-table}
\end{equation}
where $P(s|x_w)$($P(t|y_w)$) denotes the probability that a word $x_w$($y_w$) belongs to a style $s$($t$), $P(y_w|x_w)$ represents grammatic similarity of $x_w$ and $y_w$. We observe that attribute-relevance words and their proper expressions in a target attribute typically play the same grammatic role in the sentences. 
In our implementation, $P(y_w|x_w)$ is calculated with the normalized cosine similarity of word embedding \cite{Mikolov2013DistributedRO}, $P(s|x_w)$ and $P(t|y_w)$ are estimated as follows: 
\begin{equation}
\begin{aligned}
P(s|x_w) = \frac{F(s,x_w)}{F(s,x_w) + F(t,x_w)} \\
P(t|y_w) = \frac{F(t,y_w)}{F(s,y_w) + F(t,y_w)}
\end{aligned}
\label{equ:style-prob}
\end{equation}
where $F(s,x_w)$($F(t,y_w)$) represents the frequency of a word $x_w$($y_w$) appearing in datasets with attribute $s$($t$).

Specifically, as shown in the model initialization part of Figure \ref{fig:model-framework}, we learn word embeddings of all the words using source style corpus $X$ and target style corpus $Y$, based on which, the grammatic similarity model $P(y_w|x_w)$ can be learnt.
Meanwhile, with style specific corpus $X$ and $Y$, we can gain the style preference models $P(s|x_w)$ and $P(t|y_w)$.  
By incorporating these three models, we can approximate the transfer probability $P_{s \rightarrow t}(y_w|x_w) $, which is used to extract high-confidence word-level style mapping.
For instance, both ``hate" and ``love" play similar grammatical roles in the sentence, so their embeddings are very similar, and cosine-based similarity is very high. 
Additionally, ``hate'' is more inclined to appear in the negative text, while ``love" is more likely to occur in the positive text. So the two style preference probabilities are also high, which lead to a high translation probability $P_{s \rightarrow t}(``\text{love}"|``\text{hate}")$.
The inverse translation table $P_{t \rightarrow s}(y_w|x_w)$ can be generated in the same way.

To upgrade the transfer knowledge from word-level to sentence-level, we build bidirectional SMT translation systems with transfer tables and style specific language models.
Our style specific language models are based on 4-gram language models and trained using the modified Kneser-Ney smoothing algorithm over the corresponding corpus.
The features of SMT translation systems are designed as two word-level translation probabilities, two language model scores, and one word count penalty. 
For the source-to-target translation system, all the feature weights are 1, except the source style language model as -1, and similarly, the weight of target language model is set to -1 with all the remains as 1 for target-to-source translation system. 
With the SMT-based translation systems, we generate the translations of unaligned sentences $X$ and $Y$, and pair them to construct pseudo-parallel data.

\subsection{Iterative Back-Translation}

With the pseudo data generated in the first stage, we pre-train bidirectional NMT-based style transfer systems (\( P^0_{s\rightarrow t}(y|x) \) and \( P^0_{t\rightarrow s}(x|y) \)). 
In this subsection, we will start with our unsupervised training objective, based on which an iterative back-translation method is designed to further improve initial NMT-based transfer models.

Given two unaligned datasets $X=\{x_1,..., x_{n}\}$ and $Y=\{y_1,..., y_{m}\}$ labeled with attributes $s$ and $t$ respectively, the common unsupervised training objective is to maximize the likelihood of observed data:
\begin{equation}
\begin{aligned}
L^*(\theta_{s\rightarrow t},\theta_{t \rightarrow s}) = \sum_{i=1}^n\log P(x_{i}) + \sum_{i=1}^m \log P(y_{i})
\end{aligned}
\label{equ:lm-prob}
\end{equation}
where $P(x_{i})$ and $P(y_{i})$ denote the language probabilities of sentences $x_{i}$ and $y_{i}$, $\theta_{s\rightarrow t}$ and $\theta_{t \rightarrow s}$ are model parameters of $P_{s\rightarrow t}(y|x)$ and $P_{t\rightarrow s}(x|y)$ respectively.
Following \citet{zhang2018joint}'s derivation, we can get the lower bound of the training objective in Equation \ref{equ:lm-prob} as:
\begin{equation}
\begin{aligned}
L_1(\theta_{s\rightarrow t},\theta_{t \rightarrow s}) = & \sum_{i=1}^n \text{E}_{y \sim P_{s \rightarrow t}(y|x_{i})} \log P_{t \rightarrow s} (x_{i} | y) \\
 + & \sum_{i=1}^m \text{E}_{x \sim P_{t \rightarrow s}(x|y_{i})} \log P_{s \rightarrow t} (y_{i} | x)
\end{aligned}
\label{equ:lower-bound}
\end{equation}
This new training objective actually turns the unsupervised problem into a supervised one by generating pseudo sentence pairs via a back-translation method \cite{Sennrich2016ImprovingNM}, in which the first term denotes that the pseudo sentence pairs generated by the source-to-target model $P_{s \rightarrow t}(y|x)$ are used to update the target-to-source model $P_{t \rightarrow s}(x|y)$, and the second term means use of the target-to-source model $P_{t \rightarrow s}(x|y)$ to generate pseudo data for the training of the source-to-target model $P_{s \rightarrow t}(y|x)$. 
In this way, two style transfer models ($P_{s \rightarrow t}(y|x)$ and $P_{t \rightarrow s}(x|y)$) can boost each other in an iterative process, as illustrated in the iterative back-translation part of Figure \ref{fig:model-framework}.

In practice, it is intractable to calculate Equation \ref{equ:lower-bound}, since we need to sum over all candidates in an exponential search space for expectation computation.
This problem is usually alleviated by sampling \cite{Shen2016MinimumRT,Kim2016SequenceLevelKD}.
Following previous methods, the top-k translation candidates generated by beam search strategy are used for approximation.

In addition, with the weak supervision of the pseudo corpus, the learnt style transfer models are far from perfect, especially at the beginning of the iteration. 
The generated pseudo data may contains errors. 
Sometimes, the style of generated output is wrong, and such an error can be amplified in the iteration training.
To tackle this issue, we introduce an external style classifier to provide a reward to punish poor pseudo sentence pairs.
Specifically, the samples generated by $P_{s\rightarrow t}(y|x)$ or $P_{t\rightarrow s}(x|y)$ are expected to have high scores assigned by the style classifier.
The objective of this reward-based training is to maximize the expected probability of pre-trained style classifier:
\begin{equation}
\begin{aligned}
L_2(\theta_{s\rightarrow t},\theta_{t \rightarrow s}) = & \sum_{i=1}^n \text{E}_{y \sim P_{s \rightarrow t}(y|x_{i})} Q(t|y) \\
 + & \sum_{i=1}^m \text{E}_{x \sim P_{t \rightarrow s}(x|y_{i})}  Q(s|x)
\end{aligned}
\label{equ:style-classifier}
\end{equation}
where $Q(t|y)$($Q(s|x)$) denotes the probability of style $t$($s$) given the generated sentence $y$($x$).
This probability is assigned by a pre-trained style classifier and is subjected to $Q(t|.)=1-Q(s|.)$.
For the style classifier, the input sentence is encoded into a vector by a bidirectional GRU with an average pooling layer over the hidden states, and a sigmoid output layer is used to predict the classification probability. The style classifier is trained by maximum likelihood estimation (MLE) using two datasets $X$ and $Y$.

Combining Equations \ref{equ:lower-bound} and \ref{equ:style-classifier}, we get the final unsupervised training objective:
\begin{equation}
 L(\theta_{s\rightarrow t},\theta_{t \rightarrow s}) =  L_1(\theta_{s\rightarrow t},\theta_{t \rightarrow s}) + L_2(\theta_{s\rightarrow t},\theta_{t \rightarrow s}) 
\label{equ:loss-total}
\end{equation}
The partial derivative of $L(\theta_{s\rightarrow t},\theta_{t \rightarrow s})$ with respect to $\theta_{s\rightarrow t}$ and $\theta_{t \rightarrow s}$ can be written as follows:
\begin{align}
& \begin{aligned}
 \frac{\partial L(\theta_{s\rightarrow t},\theta_{t \rightarrow s})}{\partial \theta_{s\rightarrow t}} = \sum_{i=1}^m \text{E}_{x \sim P_{t \rightarrow s}(x|y_{i})} \frac{\partial \log P_{s \rightarrow t} (y_{i} | x) }{ \partial \theta_{s\rightarrow t} } \\
 + \sum_{i=1}^n \text{E}_{y \sim P_{s \rightarrow t}(y|x_{i})} [Q(t|y) \frac{\partial \log P_{s \rightarrow t} (y | x_{i}) }{ \partial \theta_{s\rightarrow t} } ]
\end{aligned}
\label{equ:gradient-s2t} \\
& \begin{aligned}
 \frac{\partial L(\theta_{s\rightarrow t},\theta_{t \rightarrow s})}{\partial \theta_{t\rightarrow s}} = \sum_{i=1}^n \text{E}_{y \sim P_{s \rightarrow t}(y|x_{i})} \frac{\partial \log P_{t \rightarrow s} (x_{i} | y) }{ \partial \theta_{t\rightarrow s} } \\
 + \sum_{i=1}^m \text{E}_{x \sim P_{t \rightarrow s}(x|y_{i})} [Q(s|x) \frac{\partial \log P_{t \rightarrow s} (x | y_{i}) }{ \partial \theta_{t \rightarrow s} } ]
\end{aligned}
\label{equ:gradient-t2s}
\end{align}
where $\frac{\partial \log P_{s \rightarrow t} (y | x_{i}) }{ \partial \theta_{s\rightarrow t} }$ and $\frac{\partial \log P_{t \rightarrow s} (x | y_{i}) }{ \partial \theta_{t \rightarrow s} }$ are the gradients specified with a standard sequence-to-sequence network.
Note that when maximizing the objective function $L_1(\theta_{s\rightarrow t},\theta_{t \rightarrow s})$, we do not back-prop through the reverse model which generates the data, following \citet{zhang2018joint} and \citet{lample2018phrase} .
The whole iterative back-translation training is summarized in Algorithm \ref{alg:iter-back-trans}.

\begin{algorithm}[t]
\caption{Iterative Back-Translation Training}
\label{alg:iter-back-trans}
\hspace*{\algorithmicindent} \textbf{Input:} Unpaired datasets $X=\{x_1,...,x_{n}\}$ and $Y=\{y_1,...,y_{m}\}$ with different attributes $s$ and $t$, initial NMT-based models \( P^0_{s\rightarrow t}(y|x) \) and \( P^0_{t\rightarrow s}(x|y) \), style classifier $Q(s|x)$($Q(t|y)$); \\
\hspace*{\algorithmicindent} \textbf{Output:} Bidirectional NMT-based style transfer models $P_{s\rightarrow t}(y|x)$ and $P_{t\rightarrow s}(x|y)$;
\begin{algorithmic}[1]
\Procedure{training process}{}
\While{$k < $ Max\_Epoches}
\State Use model $P^{k-1}_{s \rightarrow t}(y|x)$ to translate dataset $X=\{x_1,...,x_{n}\}$, yielding pseudo-parallel data \( D^{k}_{s} = \{(x_i, y^{(k)}_i)\}^{n}_{i=1} \);
\State Use model $P^{k-1}_{t \rightarrow s}(x|y)$ to translate dataset $Y=\{y_1,...,y_{m}\}$, yielding pseudo-parallel data \( D^{k}_{t} = \{(x^{(k)}_i, y_i)\}^{m}_{i=1} \);
\State Update model $P^{k}_{s \rightarrow t}(y|x)$ with Equation \ref{equ:gradient-s2t} using pseudo-parallel data \( D^{k}_{s} \), \( D^{k}_{t} \) and \( Q(t|y) \); 
\State Update model $P^{k}_{t \rightarrow s}(x|y)$ with Equation \ref{equ:gradient-t2s} using pseudo-parallel data \( D^{k}_{s} \), \( D^{k}_{t} \) and \( Q(s|x) \);
\EndWhile
\EndProcedure
\end{algorithmic}
\end{algorithm}

\section{Experiments}

\subsection{Setup}

To examine the effectiveness of our proposed approach, we conduct experiments on three datasets, including altering sentiments of Yelp reviews, altering sentiments of Amazon reviews, and altering image captions between romantic and humorous. 
Following previous work \cite{Fu2017StyleTI,Li2018DeleteRG}, we measure the accuracy of style transfer and the quality of content preservation with automatic and manual evaluations.

\paragraph{Datasets} 
To compare our work with state-of-the-art approaches, we follow the experimental setups and datasets\footnote{https://github.com/lijuncen/Sentiment-and-Style-Transfer} in \citet{Li2018DeleteRG}'s work:
\begin{itemize}
\item \textbf{Yelp:} This dataset consists of Yelp reviews. We consider reviews with a rating above three as positive samples and those below three as negative ones.
\item \textbf{Amazon:} This dataset consists of amounts of product reviews from Amazon \cite{He2016UpsAD}. Similar to Yelp, we label the reviews with a rating higher than three as positive and less than three as negative.
\item \textbf{Captions:} This dataset consists of image captions \cite{Gan2017StyleNetGA}. Each example is labeled as either romantic or humorous.
\end{itemize}
The statistics of the Yelp, Amazon and Captions datasets are shown in Table \ref{table:stn} and \ref{table:vocab}.
Additionally, \citet{Li2018DeleteRG} hire crowd-workers on Amazon Mechanical Turk to write gold output for test sets of Yelp and Amazon datasets,\footnote{The Captions dataset is actually an aligned corpus that contains captions for the same image in different styles, so we do not need to edit output for the test set of the Captions dataset. In our experiments, we also do not use these alignments.} in which workers are required to edit a sentence to change its sentiment while preserving its content.
With human reference outputs, an automatic evaluation metric, such as BLEU \cite{Papineni2002BleuAM}, can be used to evaluate how well meaning is preserved.

\begin{table}[t]
\centering
\begin{tabular}{c|c|c|r|r}
\hline
Dataset                   & Attributes & \multicolumn{1}{c|}{Train} & \multicolumn{1}{c|}{Dev} & \multicolumn{1}{c}{Test} \\ \hline
\multirow{2}{*}{Yelp}     & Negative   & 180K                       & 2000                     & 500                       \\ \cline{2-5} 
                          & Positive   & 270K                       & 2000                     & 500                       \\ \hline
\multirow{2}{*}{Amazon}   & Negative   & 278K                       & 1015                     & 500                       \\ \cline{2-5} 
                          & Positive   & 277K                       & 985                      & 500                       \\ \hline
\multirow{2}{*}{Captions} & Humorous   & 6000                       & 300                      & 300                       \\ \cline{2-5} 
                          & Romantic   & 6000                       & 300                      & 300                       \\ \hline
\end{tabular}
\caption{Sentence count in different datasets.}
\label{table:stn}
\end{table}
\begin{table}[t]
\centering
\begin{tabular}{c|c|c|c}
\hline
Dataset   & Yelp & Amazon & Captions \\ \hline
Vocabulary & 10K  & 20K    & 8K       \\ \hline
\end{tabular}
\caption{Vocabulary size of different datasets.}
\label{table:vocab}
\end{table}

\begin{table*}[t]
\centering
\begin{tabular}{l|c|c|c|c|c|c}
\hline
\multirow{2}{*}{} & \multicolumn{2}{c|}{Yelp} & \multicolumn{2}{c|}{Amazon} & \multicolumn{2}{c}{Captions} \\ \cline{2-7} 
                  & Classifier     & BLEU     & Classifier      & BLEU      & Classifier       & BLEU       \\ \hline \hline
CrossAligned      & 73.2\%         & 9.06     & 71.4\%          & 1.90      & 79.1\%           & 1.82       \\ \hline
MultiDecoder      & 47.0\%         & 14.54    & 66.4\%          & 9.07      & 66.8\%           & 6.64       \\ \hline
StyleEmbedding    & 7.6\%          & 21.06    & 40.3\%          & 15.05     & 54.3\%           & 8.80       \\ \hline
TemplateBased     & 80.3\%         & 22.62    & 66.4\%          & 33.57     & 87.8\%           & \textbf{19.18}      \\ \hline
Del-Retr-Gen        & 89.8\%         & 16.00    & 50.4\%          & 29.27     & 95.8\%           & 11.98      \\ \hline
Our Approach      & \textbf{96.6\%}         & \textbf{22.79}    & \textbf{84.1\%}          & \textbf{33.90}     & \textbf{99.5\%}           & 12.69      \\ \hline
\end{tabular}
\vspace{-5pt}
\caption{Automatic evaluation results on Yelp, Amazon and Captions datasets. ``Classifier" shows the accuracy of sentences labeled by the pre-trained style classifier. ``BLEU(\%)" measures content similarity between the output and the human reference.}
\label{table:auto-eval}
\end{table*}

\begin{table*}[t]
\centering
\begin{tabular}{l|c|c|c|c|c|c|c|c|c|c|c|c}
\hline
\multirow{2}{*}{}        & \multicolumn{4}{c|}{Yelp} & \multicolumn{4}{c|}{Amazon} & \multicolumn{4}{c}{Captions} \\ \cline{2-13} 
                         & Att  & Con  & Gra  & Suc  & Att   & Con  & Gra  & Suc   & Att   & Con   & Gra   & Suc   \\ \hline \hline
CrossAligned             & 3.1  & 2.7  & 3.2  & 10\% & 2.4   & 1.8  & 3.4  & 6\%   &     3.0  &  2.2     &  3.7     &  14\%     \\ \hline
MultiDecoder              & 2.4  & 3.1  & 3.2  & 8\% &  2.4  & 2.3  & 3.2  & 7\%  &  2.8    & 3.0      &  3.4     &   16\%    \\ \hline
StyleEmbedding            & 1.9   & 3.5   & 3.3  & 7\%  & 2.2  & 2.9  & 3.4  & 10\%  &   2.7    &  3.2    &  3.3    &  16\%     \\ \hline
TemplateBased            & 2.9  & 3.6  & 3.1  & 17\% & 2.1   & 3.5  & 3.2  & 14\%  &    3.3   &  \textbf{3.8}    &   3.3    &   23\%    \\ \hline
Del-Retr-Gen & 3.2  & 3.3  & 3.4  & 23\% & 2.7   & \textbf{3.7}  & 3.8  & 22\%  &    3.5   &   3.4    &  \textbf{3.8}     &  32\%     \\ \hline
Our Approach             & \textbf{3.5}  & \textbf{3.7}  & \textbf{3.6}  & \textbf{33\%} & \textbf{3.3}   & \textbf{3.7}  & \textbf{3.9}  & \textbf{30\%}  &   \textbf{3.6}    &   \textbf{3.8}    &  3.7     &  \textbf{37\%}     \\ \hline
\end{tabular}
\vspace{-5pt}
\caption{Human evaluation results on Yelp, Amazon and Captions datasets. We show average human ratings for style transfer accuracy (Att), preservation of meaning (Con), fluency of sentences (Gra) on a 1 to 5 Likert scale. "Suc" denotes the overall success rate. We consider a generated output "successful" if it is rated 4 or 5 on all three criteria (Att, Con, Gra).}
\end{table*}

\paragraph{Baselines} 
We compare our approach with five state-of-the-art baselines: \textbf{CrossAligned} \cite{Shen2017StyleTF}, \textbf{MultiDecoder} \cite{Fu2017StyleTI}, \textbf{StyleEmbedding} \cite{Fu2017StyleTI}, \textbf{TemplateBased} \cite{Li2018DeleteRG} and  \textbf{Del-Retr-Gen} (Delete-Retrieve-Generate) \cite{Li2018DeleteRG}.
The former three methods are based on auto-encoder neural networks and leverage an adversarial framework to help systems separate style and content information. 
TemplateBased is a retrieve-based method that first identifies attribute-relevance words and then replaces them with target attribute expressions, which are extracted from a similar content sentence retrieved from the target style corpus.  
Del-Retr-Gen is a mixed model combining the TemplateBased method and an RNN-based generator, in which the RNN-based generator produces the final output sentence based on the content and the extracted target attributes.

\paragraph{Training Details}
For the SMT model in our approach, we use Moses\footnote{https://github.com/moses-smt/mosesdecoder} with a translation table initialized as described in the Model Initialization Section.
The language model is a default smoothed n-gram language model and the reordering model is disabled.
The hyper-parameters of different SMT features are assigned as described in the Model Initialization Section.

RNNSearch \cite{Bahdanau2014NeuralMT} is adopted as the NMT model in our approach, which uses a single layer GRU for the encoder and decoder networks, enhanced with a feed-forward attention network.
The dimension of word embedding (for both source and target words) and hidden layer are set to 300.
All parameters are initialized using a normal distribution with a mean of 0 and a variance of $\sqrt{6/(d_{row}+d_{col})}$, where $d_{row}$ and $d_{col}$ are the number of rows and columns of the parameter matrix \cite{glorot2010understanding}.
Each model is optimized using the Adadelta \cite{zeiler2012adadelta} algorithm with a mini-batch size 32.
All of the gradients are re-normalized if the norm exceeds 2. 
For the training iteration in Algorithm \ref{alg:iter-back-trans}, best 4 samples generated by beam search strategy are used for training, and we run 3 epochs for Yelp and Amazon datasets, 30 epochs for the Captions dataset.  
At test time, beam search is employed to find the best candidate with a beam size 12.

\subsection{Automatic Evaluation}

In automatic evaluation, following previous work \cite{Shen2017StyleTF,Li2018DeleteRG}, we measure the accuracy of style transfer for the generated sentences using a pre-trained style classifier,\footnote{We train another style classifier for our iterative back-translation training process.} and adopt a case-insensitive BLEU metric to evaluate the preservation of content.
The BLEU score is computed using Moses \textit{multi-bleu.perl} script.
For each dataset, we train the style classifier on the same training data.

Table \ref{table:auto-eval} shows the automatic evaluation results of different models on Yelp, Amazon and Captions datasets.
We can see that CrossAligned obtains high style transfer accuracy but sacrifices content consistency. 
In addition, MultiDecoder and StyleEmedding can help the preservation of content but reduce the accuracy of style transfer.
Compared with previous methods, TemplateBased and Del-Retr-Gen achieve a better balance between the transfer accuracy and the content preservation.
Our approach achieves significant improvements over CrossAligned, MultiDecoder, StyleEmedding and Del-Retr-Gen in both transfer accuracy and quality of content preservation.

Compared with TemplateBased, our method achieves much better accuracy of style transfer, but with a lower BLEU score on the Captions dataset. 
The reason is that there are more different expressions to exhibit romantic and humorous compared with changing sentiment.
A BLEU score based on a single human reference cannot precisely measure content consistency.
In addition, as argued in \citet{Li2018DeleteRG}, the BLEU metric, which is lack of automatic fluency evaluation, favors systems like TemplateBased, which only replaces a few words in the sentence.
However, grammatical mistakes are easily made when replacing with inappropriate words.
In order to reflect grammatical mistakes in the generated sentence, we conduct human evaluation with fluency as one of the criteria.

\begin{table*}[t]
\centering
\begin{tabular}{l|c|c|c|c|c|c}
\hline
\multirow{2}{*}{Models} & \multicolumn{2}{c|}{Yelp} & \multicolumn{2}{c|}{Amazon} & \multicolumn{2}{c}{Captions} \\ \cline{2-7} 
                  & Classifier     & BLEU     & Classifier      & BLEU      & Classifier       & BLEU       \\ \hline \hline
SMT-based      & 94.6\%         & 8.82     & 81.2\%          & 7.46      & 82.8\%           & 5.04      \\ \hline
NMT-based (Iteration 0)      & 70.5\%         & 15.81    & 68.4\%          & 16.36      & 66.5\%           & 8.72       \\ \hline
NMT-based    &\textbf{96.6}\%         & 22.79    & \textbf{84.1}\%          & 33.90     & \textbf{99.5}\%           & 12.69       \\ \hline
NMT-based (w/o Style Classifier)    & 80.4\%         & \textbf{24.48}    & 75.6\%          & \textbf{35.34}     & 79.3\%           & \textbf{13.93}      \\ \hline
\end{tabular}
\vspace{-5pt}
\caption{Automatic evaluation results of each component of our approach on Yelp, Amazon andf Captions datasets.}
\label{table:analysis}
\end{table*}

\subsection{Human Evaluation}

While automatic evaluation provides an indication of style transfer quality, it can not evaluate the quality of transferred text accurately.
To further verify the effectiveness of our approach, we perform a human evaluation on the test set.
For each dataset, we randomly select 200 samples for the human evaluation (100 for each attribute).
Each sample contains the transformed sentences generated by different systems given the same source sentence.
Then samples are distributed to annotators in a double-blind manner.\footnote{We distribute each sample to 5 native speakers and use Fleiss’s kappa to judge agreement among them. The Fleiss’s kappa score is 0.791 for the Yelp dataset, 0.763 for the Amazon dataset and 0.721 for the Caption dataset.}
Annotators are asked to rate each output for three criteria on a likert scale from 1 to 5: style transfer accuracy (Att), preservation of content (Con) and fluency of sentences (Gra).
Finally, the generated sentence is treated as ``successful" when it is scored 4 or 5 on all three criteria.  

Table \ref{table:auto-eval} shows the human evaluation results. 
It can be clearly observed that our proposed method achieves the best performance among all systems, with 10\%, 8\% and 5\% point improvements than Del-Retr-Gen on Yelp, Amazon and Captions respectively, which demonstrates the effectiveness of our proposed method.
Compared with Del-Retr-Gen, our method is rated higher on all three criteria.

By comparing two evaluation results, we find that there is a positive correlation between human evaluation and automatic evaluation in terms of both accuracy of style transfer and preservation of content. 
This indicates the usefulness of automatic evaluation metrics in model development.
However, since current automatic evaluation cannot evaluate the quality of transferred text accurately, the human evaluation is necessary and accurate automatic evaluation metrics are expected in the future.

\subsection{Analysis}

We further investigate the contribution of each component of our method during the training process.
Table \ref{table:analysis} shows automatic evaluation results of SMT-based and NMT-based transfer systems in our approach on the Yelp, Amazon and Captions datasets.
We find that, with the help of word-level translations table and style specific language models, the SMT-based transfer model gains high accuracy in style transfer but fails in preserving content.
Given pseudo data generated by the SMT-based model, the NMT-based model (Iteration 0) can better integrate the translation and language models, resulting in sentences with better content consistency. 
Using our iterative back-translation algorithm, the pre-trained NMT-based model can then be significantly improved in terms of both accuracy of style transfer and preservation of content.
This result proves that the iterative back-translation algorithm can effectively leverage unaligned sentences.
Besides, the style classifier plays a key role to guarantee the success of style transfer, without which, the transfer accuracy of the NMT-based model is obviously declining due to imperfect pseudo data.
We also show some system outputs in Table \ref{table:examples}.

\section{Related Work}

Language style transfer without a parallel text corpus has attracted more and more attention due to recent advances in text generation tasks.
Many approaches have been proposed to build style transfer systems and achieve promising performance \cite{Hu2017TowardCG,Shen2017StyleTF,Fu2017StyleTI,Li2018DeleteRG,Prabhumoye2018StyleTT}. 
\citet{Hu2017TowardCG} leverage an attribute classifier to guide the generator to produce sentences with desired attribute (e.g. sentiment, tense) in the Variational Auto-encoder (VAE) framework.
\citet{Shen2017StyleTF} first supply a theoretical analysis of language style transfer using non-parallel text. 
They propose a cross-aligned auto-encoder with discriminator architecture, in which an adversarial discriminator is used to align different styles.

Instead of only considering the style transfer accuracy as in previous work, \citet{Fu2017StyleTI} introduce the content preservation as another evaluation metric and design two models, which encode the sentence into latent content representation and leverage the adversarial network to separate style and content information.
Similarly, \citet{Prabhumoye2018StyleTT} attempt to use external NMT models to rephrase the sentence and weaken the effect of style attributes, based on which, multiple decoders can better preserve sentence meaning when transferring style.
More recently, \citet{Li2018DeleteRG} design a delete-retrieve-generate system which hybrids the retrieval system and the neural-based text generation model. 
They first identify attribute-related words and remove them from the input sentence.
Then the modified input sentence is used to retrieve a similar content sentence from the target style corpus, based on which corresponding target style expressions are extracted to produce the final output with an RNN-based generator.

Different from previous methods, we treat language style transfer as a special MT task where the source language is in one style and the target language in another style. 
Based on this, we adopt an attention-based sequence-to-sequence model to transform the style of a sentence.
Further, following the key framework of unsupervised MT methods \cite{Artetxe2018UnsupervisedNM,Piotr2018UnsupervisedMT,lample2018phrase} to deal with the problem of lacking parallel corpus, a two-stage joint training method is proposed to leverage unpaired datasets with attribute information.

However, there are two major differences between our proposed approach and the existing unsupervised NMT method: 1) Building the word-to-word translation system in unsupervised NMT relies on the homogeneity of cross-lingual word embedding space, which is impossible for a style transfer whose input and output are in the same language. To deal with that, we propose a new method taking advantage of style-preference information and word embedding similarity to build the word-to-word transfer system; 2) We leverage the style classifier to filter the bad generated pseudo sentences, and its score is used as rewards to stabilize model training. Table \ref{table:analysis} shows that introducing a style classifier can better guarantee the transferred style. In summary, an unsupervised NMT method cannot be directly applied in style transfer tasks, and we modify two important components to make it work.

\section{Conclusion}

In this paper, we have presented a two-stage joint training method to boost source-to-target and target-to-source style transfer systems using non-parallel text.
In the first stage, we build bidirectional word-to-word style transfer systems in a SMT framework to generate pseudo sentence pairs, based on which, two initial NMT-based style transfer models are constructed.  
Then an iterative back-translation algorithm is employed to better leverage non-parallel text to jointly improve bidirectional NMT-based style transfer systems.
Empirical evaluations are conducted on Yelp, Amazon and Captions datasets, demonstrating that our approach outperforms previous state-of-the-art models in terms of both accuracy of style transfer and quality of input-output correspondence.

In the future, we plan to further investigate the use of our method on other style transfer tasks.
In addition, we are interested in designing more accurate and complete automatic evaluation for this task.

\bibliographystyle{aaai}
\bibliography{aaai19}

\appendix

\begin{table*}[t]
\centering
\begin{tabular}{ll}
\hline 
\multicolumn{2}{c}{From negative to positive (Yelp)}                                                                                 \\ \hline
\multicolumn{1}{l|}{Source}         & the food was so-so and very over priced for what you get .                                     \\ \hline \hline
\multicolumn{1}{l|}{CrossAligned}   & the food was fantastic and very very nice for what you .                                       \\
\multicolumn{1}{l|}{MultiDecoder}   & the food was low up and over great , see you need .                                            \\
\multicolumn{1}{l|}{StyleEmbedding} & the food was so-so and very over priced for what you get .                                     \\
\multicolumn{1}{l|}{TemplateBased}  & the food was so-so and very over priced for what you get just right .                          \\
\multicolumn{1}{l|}{Del-Retr-Gen}   & the service is fantastic and the food was so-so and the food is very priced for what you get . \\
\multicolumn{1}{l|}{Our Approach}   & the food was decent and very perfectly priced for what you get .                               \\ \hline  
\multicolumn{2}{c}{From positive to negative (Yelp)}                                                                                 \\ \hline 
\multicolumn{1}{l|}{Source}         & the service was top notch and the food was a bit of heaven .                                   \\ \hline \hline
\multicolumn{1}{l|}{CrossAligned}   & the service was top notch and the room was very expensive on me .                              \\
\multicolumn{1}{l|}{MultiDecoder}   & the service was top and the food was a bit of this plate .                                     \\
\multicolumn{1}{l|}{StyleEmbedding} & the service was top notch and the food was a bit of heaven .                                   \\
\multicolumn{1}{l|}{TemplateBased}  & slow the food was a bit of                                                                     \\
\multicolumn{1}{l|}{Del-Retr-Gen}   & the food was a bit of weird .                                                                  \\
\multicolumn{1}{l|}{Our Approach}   & the service was lacking and the food was a bit of sick .                                       \\ \hline
\multicolumn{2}{c}{From negative to positive (Amazon)}                                                                               \\ \hline  
\multicolumn{1}{l|}{Source}         & this is not worth the money and the brand name is misleading .                                 \\ \hline \hline
\multicolumn{1}{l|}{CrossAligned}   & this is not the best and the best is not great .                                               \\
\multicolumn{1}{l|}{MultiDecoder}   & this is not worth the money and this pan , at amazon .                                         \\
\multicolumn{1}{l|}{StyleEmbedding} & this is not worth the money and the brand name is the price .                                  \\
\multicolumn{1}{l|}{TemplateBased}  & you can not beat the price and the brand name is misleading .                                    \\
\multicolumn{1}{l|}{Del-Retr-Gen}   & well worth the money and the brand name is misleading .                                        \\
\multicolumn{1}{l|}{Our Approach}   & this is definitely worth the money and the brand name is illustrated .                         \\ \hline
\multicolumn{2}{c}{From positive to negative (Amazon)}                                                                               \\ \hline 
\multicolumn{1}{l|}{Source}         & i would definitely recommend this for a cute case .                                            \\ \hline  \hline
\multicolumn{1}{l|}{CrossAligned}   & i would not recommend this for a long time .                                                   \\
\multicolumn{1}{l|}{MultiDecoder}   & i would definitely recommend this for a bra does it .                                          \\
\multicolumn{1}{l|}{StyleEmbedding} & i would definitely recommend this for a cute case .                                            \\
\multicolumn{1}{l|}{TemplateBased}  & skip this one for a cute case .                                                                \\
\multicolumn{1}{l|}{Del-Retr-Gen}   & i would not recommend this for a cute case .                                                   \\
\multicolumn{1}{l|}{Our Approach}   & i would definitely not recommend this for a cute case .                                        \\ \hline
\multicolumn{2}{c}{From factual to romantic (Captions)}                                                                              \\ \hline  
\multicolumn{1}{l|}{Source}         & a man and woman against a pink background smile .                                              \\ \hline \hline
\multicolumn{1}{l|}{CrossAligned}   & a man in a red shirt is running on a beach .                                                   \\
\multicolumn{1}{l|}{MultiDecoder}   & a man and woman on a red crowd looks .                                                         \\
\multicolumn{1}{l|}{StyleEmbedding} & a man and woman watch a people play music .                                                    \\
\multicolumn{1}{l|}{TemplateBased}  & a man and woman against a pink background smile loved .                                        \\
\multicolumn{1}{l|}{Del-Retr-Gen}   & a man and woman watches a pink street to show his lover .                                      \\
\multicolumn{1}{l|}{Our Approach}   & a man and woman crossing a kiss together dreaming of love .                                    \\ \hline
\multicolumn{2}{c}{From factual to humorous (Captions)}                                                                              \\ \hline  
\multicolumn{1}{l|}{Source}         & a young man dances by a fountain .                                                             \\ \hline \hline
\multicolumn{1}{l|}{CrossAligned}   & a man is running on a beach to find the space .                                                \\
\multicolumn{1}{l|}{MultiDecoder}   & a young man stands next like a car .                                                           \\
\multicolumn{1}{l|}{StyleEmbedding} & a young man dances along an inflatable fountain .                                              \\
\multicolumn{1}{l|}{TemplateBased}  & a young man dances by a fountain deadly .                                                      \\
\multicolumn{1}{l|}{Del-Retr-Gen}   & a young man is running off for supremacy .                                                     \\
\multicolumn{1}{l|}{Our Approach}   & a young man sits by a fountain like a monkey with a smiley face .                              \\ \hline
\end{tabular}
\caption{Style transfer examples of different systems on Yelp, Amazon and Captions datasets.}
\label{table:examples}
\end{table*}

\end{document}